\documentclass[10pt,twocolumn]{article}

\usepackage[margin=0.75in]{geometry}
\usepackage{amsmath,amssymb,amsfonts}
\usepackage{graphicx}
\usepackage{booktabs}
\usepackage{cite}
\usepackage{hyperref}
\usepackage{caption}
\usepackage{textcomp}
\usepackage{cite}
\usepackage{algorithm}
\usepackage{algpseudocode}
\usepackage{multicol}
\usepackage{multirow}

\title{A study on a Real-Time VR-Based Teleoperation Framework for Manipulator in Dynamic Environment}

\author{
    InGyu Choi$^{1}$, GeonYeong Go$^{2}$, SunWoo Ahn$^{2}$, HyoJae Kang$^{3}$, and Min-Sung Kang$^{4,*}$ \\
    $^{1}$Department of Robotics, Hanyang University \\
    $^{2}$Department of Smart Construction Engineering, Hanyang University \\
    $^{3}$Department of Interdisciplinary Robot Engineering Systems, Hanyang University \\
    $^{4}$School of Smart Convergence Engineering, Hanyang University, Ansan \\
    $^{*}$Corresponding author
}

\begin{document}
\date{}
\maketitle

\thispagestyle{plain}
\footnotetext{This manuscript has been submitted for possible publication.\\
This work was supported by the Technology Innovation Program (RS-2024-00442513, Development of robotic manipulation task learning based on foundation model to understand and reason about task situations) funded by the Ministry of Trade Industry \& Energy (MOTIE, Korea).}

\begin{abstract}
Robot teleoperation enables safe, non-contact task execution in hazardous environments where direct human access is difficult, and its application has expanded with recent VR technologies. Many VR teleoperation studies, however, have primarily served as data-collection tools for robot imitation learning, so they often do not explicitly address dynamic obstacles, workspace changes, or collision risks during operation. For real deployment aimed at operator safety, teleoperation must react to dynamic situations with low latency and remain robust to mistakes made by inexperienced operators. This paper presents a VR teleoperation framework that supports real-time manipulation while handling collisions with both static and moving obstacles. The framework integrates GPU-accelerated inverse kinematics and trajectory optimization within a VR interface to generate feasible joint commands at each control cycle under robot constraints. Experiments with a 7-DoF manipulator demonstrate stable online behavior and collision-aware motion generation across three scenarios: obstacle-free, static-obstacle, and moving-obstacle environments. The results indicate that the proposed approach generates motion consistent with the operator’s command while producing safe detours when obstacles interfere with the commanded path.

\end{abstract}

\section{Introduction}

Teleoperation is a practical solution for operating robotic manipulators in hazardous or hard-to-access environments where direct human intervention is unsafe or infeasible, such as disaster response, construction sites, and high-risk industrial facilities \cite{c1,c2}. Traditionally, industrial robots have been deployed in highly structured environments and controlled through pre-taught waypoints or precomputed trajectories, achieving high efficiency and precision for repetitive tasks. However, as robots are increasingly expected to replace human labor in unstructured and dynamic environments, these conventional control strategies become difficult to apply due to frequent changes in workspace conditions and unexpected obstacles. With the recent adoption of virtual reality (VR) interfaces, ﻿teleoperation systems have evolved to provide more intuitive six-degree-of-freedom (6-DoF) interaction and enhanced situational awareness compared to conventional interfaces. VR-based teleoperation allows operators to directly manipulate the robot’s end-effector pose in a virtual environment, decoupling the operator from the physical workspace while preserving intuitive control. For real-world deployment, real-time responsiveness and operational safety are critical, since delays or unsafe motions can directly lead to task failure, collisions, or increased operator workload.

Despite recent advances in VR-based teleoperation, many existing systems still generate robot motion primarily based on end-effector commands without sufficiently considering real-time changes in the surrounding environment. Consequently, dynamic obstacles and workspace variations are not adequately reflected during motion generation, limiting the robot’s ability to safely and effectively operate in cluttered and changing scenes. In addition, as manipulators become increasingly kinematically redundant and safety requirements expand to include joint limits, collision avoidance, and manipulability constraints, the computational burden of real-time motion generation increases substantially. This often introduces communication and computation latency, causing input–response mismatch between the operator and the robot, which degrades controllability and increases cognitive load.

To address these challenges, this paper proposes a VR-based teleoperation framework that explicitly integrates real-time environment perception with GPU-accelerated optimization-based control. The workspace is continuously reconstructed online to maintain an up-to-date representation of surrounding obstacles. Based on this representation, a GPU-accelerated trajectory optimization module generates feasible joint commands at each control cycle while satisfying robot constraints such as joint limits and collision avoidance. By directly reflecting environment-derived constraints in the motion generation process, the proposed framework enables safe and responsive robot behavior under dynamic environmental changes while remaining consistent with the operator’s commanded motion.

The main contribution of this work is the presentation of a tightly integrated teleoperation architecture in which operator input, online environment perception, and constraint-aware motion optimization are unified within the same real-time control loop. The proposed framework enables robot motion to be generated not only from operator commands but also from continuously updated environmental information, thereby supporting safe and responsive manipulation in dynamic and cluttered environments.

\section{Related Works}

Teleoperation combined with eXtended Reality (XR) can be broadly categorized into Augmented Reality (AR) and Virtual Reality (VR). AR-based teleoperation overlays virtual information onto the real environment, providing visual cues that facilitate robot manipulation; however, the operator must still remain physically present in the workspace \cite{c3,c4}. In contrast, VR-based teleoperation projects images of the robot’s workspace, captured by on-site cameras, into a fully virtual environment, enabling a higher level of immersion while completely separating the operator from the robot’s physical workspace \cite{c5,c6}. Owing to this separation, VR-based teleoperation offers both intuitive interaction and improved operator safety \cite{c7}.
 
Building on these advantages, a wide range of studies have explored VR-based teleoperation. For example, VR interfaces have been used to remotely reprogram industrial robots, demonstrating that flexible robot operation is possible even without the operator’s physical presence \cite{c8}. However, such approaches typically assume static environments and local network conditions and thus do not sufficiently address dynamic environmental changes or strict real-time constraints. Other studies have focused on reducing communication latency by integrating VR teleoperation with protocols such as MQTT \cite{c6}, achieving low latency even in multi-robot scenarios. Nevertheless, these methods still do not explicitly consider dynamic obstacles or incorporate constraints such as joint limits, collision avoidance, and manipulability into the control framework.

To deploy VR-based teleoperation in dynamic environments, such as real industrial sites, robot motion must reflect not only the operator’s commands but also real-time changes in the positions of obstacles and objects. This requirement places strict real-time demands on the entire pipeline, including sensor processing, IK, collision checking, and network communication. In particular, IK computation with constraints and communication latency are dominant factors that determine overall system responsiveness.

Several studies have attempted to mitigate latency introduced by IK computation. Closed-Loop Inverse Kinematics (CLIK) has been applied to real-time arm control using Jacobian-based linearization, achieving stable tracking with bounded latency by incorporating null-space compensation near singular configurations \cite{c9,c10}. In TeleopLab, a smartphone-based teleoperation framework, TRAC-IK is employed to rapidly compute joint configurations from user-specified end-effector poses by running Jacobian-based and optimization-based solvers in parallel \cite{c11,c12}. 

\section{Methodology}
\subsection{System Overview}

\begin{figure*}[!t]
    \centering
    \includegraphics[width=0.95\textwidth]{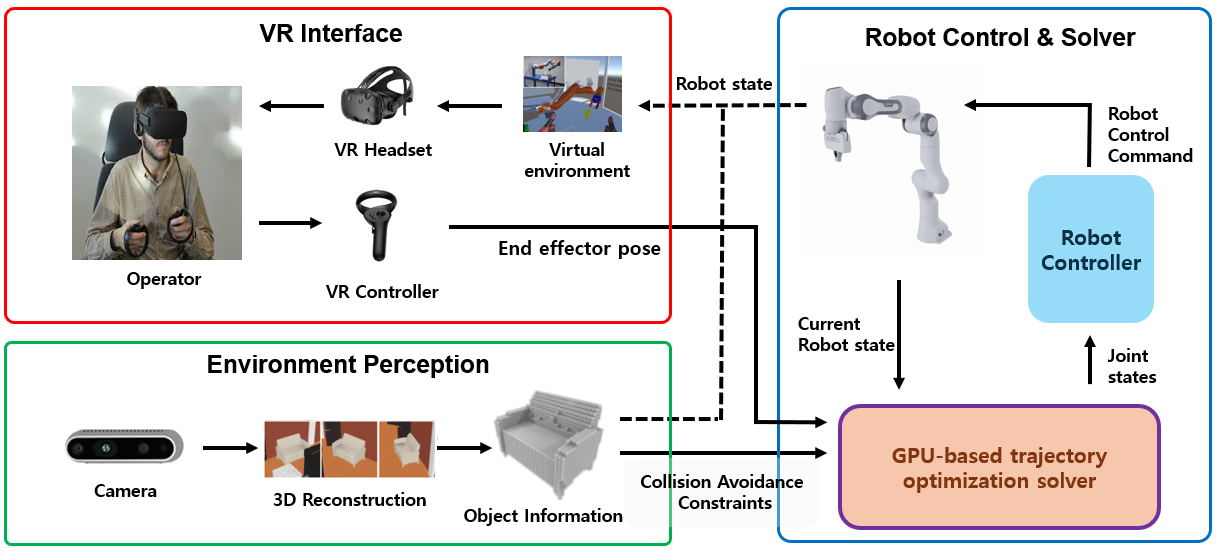}
    \caption{Overall system architecture of the proposed VR teleoperation framework}
    \label{fig1}
\end{figure*}

The proposed framework is designed to enable stable VR-based teleoperation of a manipulator in dynamic environments by tightly coupling user input, real-time environment perception, and constraint-aware command generation. The overall system consists of three main modules, as shown in Fig. ~\ref{fig1}: a VR interface module, a 3D reconstruction–based perception module, and a GPU-based optimization control module. Each module operates continuously, and their outputs are synchronized to produce feasible joint commands at every control cycle.

The VR interface module captures the operator’s 6-DoF controller motion in the virtual environment and converts it into a target end-effector pose. This target pose reflects the operator’s command and is updated online during teleoperation. The perception module maintains an up-to-date representation of the workspace using online 3D reconstruction from RGB-D observations. From the reconstructed scene, obstacle candidates are extracted and converted into a compact representation suitable for real-time control. In addition, the module removes the robot’s own geometry from the reconstruction to prevent robot links from being misinterpreted as external obstacles. The optimization control module generates joint-level commands by solving a constraint-aware motion optimization problem at each control cycle. The objective is to follow the target end-effector pose while satisfying robot constraints such as joint limits, motion bounds, and collision avoidance. Obstacle information provided by the perception module is injected into the optimization as collision-related costs and/or constraints, allowing the controller to reshape the robot trajectory online when the environment changes.

During each control iteration, the system follows a consistent pipeline: the VR module updates the target end-effector pose, the perception module updates obstacle representations from the latest reconstruction, and the optimization module computes the next feasible command based on both the target and the updated environment constraints. This closed-loop integration enables the robot to generate motion that remains consistent with the operator’s command while continuously adapting to changes in the environment.

\subsection{VR Input to Robot Target Pose Mapping}

\begin{figure*}[!t]
    \centering
    \includegraphics[width=0.95\textwidth]{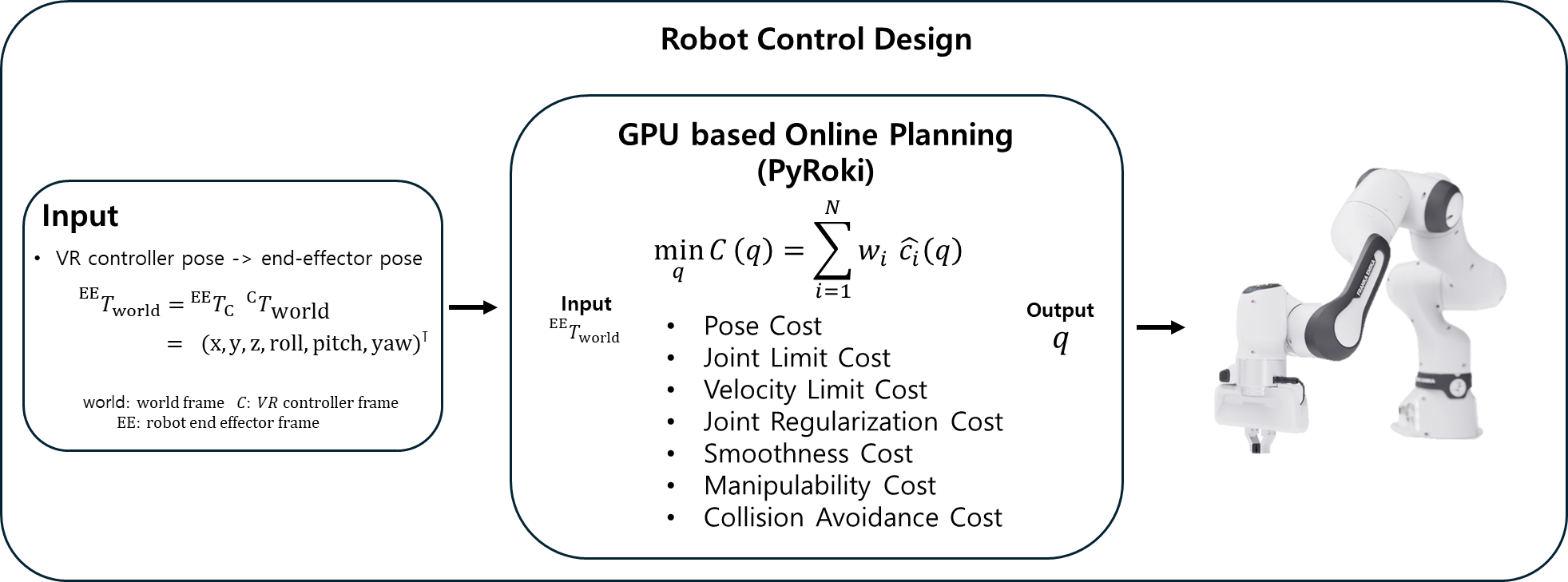}
    \caption{Block diagram of the proposed robot control design.}
    \label{fig2}
\end{figure*}

The pose of the VR controller is transformed into the target end-effector pose and used as control input (Fig. ~\ref{fig2}). The pose measured in the controller frame (C) is first expressed in the world frame (world) of the robot workspace and then mapped to the end-effector frame (EE) as

\begin{equation}
    ^{EE}T_{world} = ^{EE}T_C ^{C}T_{world} = (x,y,z,roll,pitch,yaw)^T,
\end{equation}

\noindent where $^C T_{world}$ denotes the controller pose in the world frame and $^{EE}T_C$ is a fixed transform that aligns the VR controller motion with the robot end-effector motion. The resulting $^{EE}T_{world}$ is used as the target pose for the manipulator, and the online optimization-based controller computes the corresponding joint configuration. This formulation provides an intuitive mapping from hand motion in VR to the robot end-effector and resolves inconsistencies induced by different device coordinate frames.
 
\subsection{Optimization-Based Control}

In VR-based teleoperation, the operator’s hand motion must be reflected in the end-effector motion at each control step. To achieve this while satisfying various physical constraints in real time, PyRoki \cite{c13} is employed as a GPU-based optimization framework that integrates inverse kinematics and trajectory optimization into a single problem. PyRoki simultaneously handles pose tracking, joint constraints, and collision avoidance, and computes solutions with millisecond-level latency through parallel optimization on the GPU. The robot state $q\in \mathbb{R}^n$ is determined by minimizing a weighted sum of multiple cost terms:

\begin{equation}
    \min_q C(q)=\sum_{i=1}^N w_i \hat{c}_i(q),
\end{equation}

\noindent where each term $\hat{c}_i(q)$ represents a physical objective (pose tracking, joint limits, collision avoidance, manipulability, smoothness, etc.), and $w_i$ is its associated weight. In this work, seven cost functions are used: pose tracking, joint limits, velocity limits, posture regularization, trajectory smoothness, manipulability, and collision avoidance.

\subsection{Dynamic Environment Perception}

\begin{figure}[!t]
    \centering
    \includegraphics[width=0.95\columnwidth]{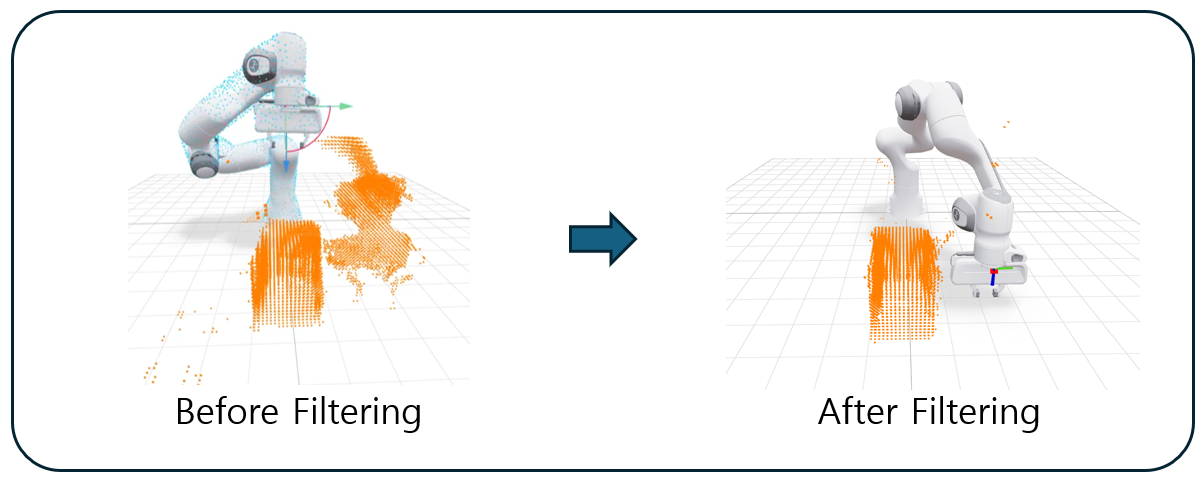}
    \caption{Overview of the perception-to-obstacle representation pipeline}
    \label{fig3}
\end{figure}

\begin{figure}[!t]
    \centering
    \includegraphics[width=0.95\columnwidth]{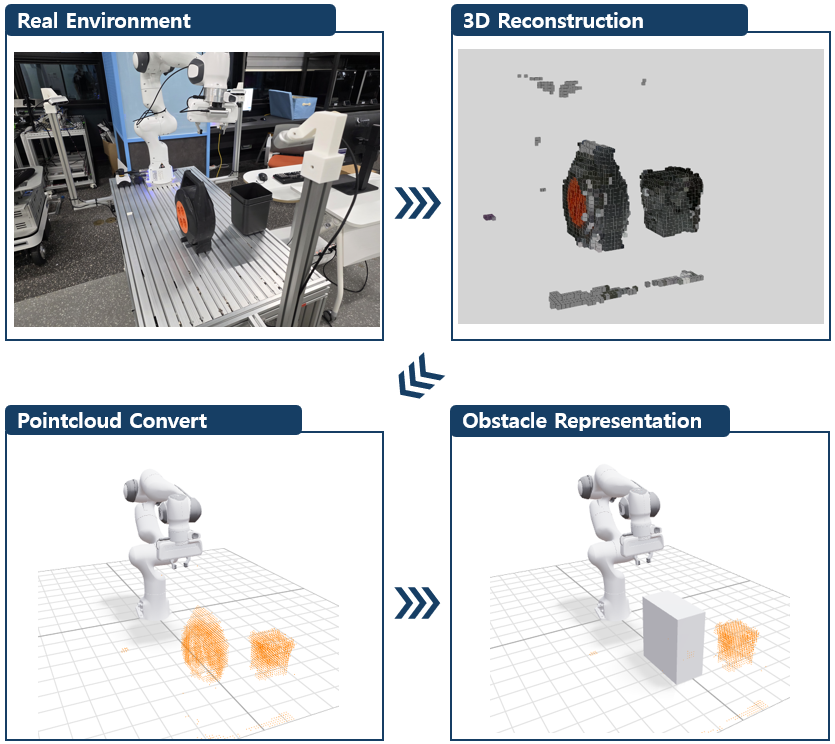}
    \caption{Robot occupancy exclusion in the reconstructed environment point cloud}
    \label{fig4}
\end{figure}

This section describes the real-time perception and reconstruction of the environment and how the resulting information is used as collision-avoidance constraints. For safe VR-based teleoperation in dynamic environments, the workspace around the robot must be reconstructed continuously, and obstacle information must be provided to the controller in real time. As illustrated in Fig. ~\ref{fig3}, RGB–D images of the workspace are first captured and fused into a voxel-based 3D reconstruction using the Nvblox \cite{c14}. To prevent the robot itself from being misinterpreted as an external obstacle, the occupied volume of the robot is explicitly removed from the reconstructed scene using the known robot geometry and kinematic model. This robot occupancy filtering step ensures that only external objects and environmental structures are preserved in the environment representation (Fig. ~\ref{fig4}). The filtered reconstruction is then converted into a point cloud in the robot frame, and DBSCAN\cite{c15}-based clustering is applied to extract individual obstacle regions. Each cluster is approximated by simple geometric primitives, which are incorporated into the motion optimization as collision-avoidance constraints, enabling safe and responsive trajectory generation in dynamic environments.

\section{System Implementation and Experimental Setup}

\subsection{System Implementation}
In this chapter, the system configuration used to realize the proposed VR-based teleoperation framework is described. The chapter is organized into two parts: Hardware Configuration, which summarizes the robot, VR devices, sensing setup, and computing platform, and Software Architecture and Communication, which explains the modular software pipeline and the dual communication scheme that together support real-time operation in dynamic environments.

\subsubsection{Hardware Configuration}

\begin{figure}[!t]
    \centering
    \includegraphics[width=0.95\columnwidth]{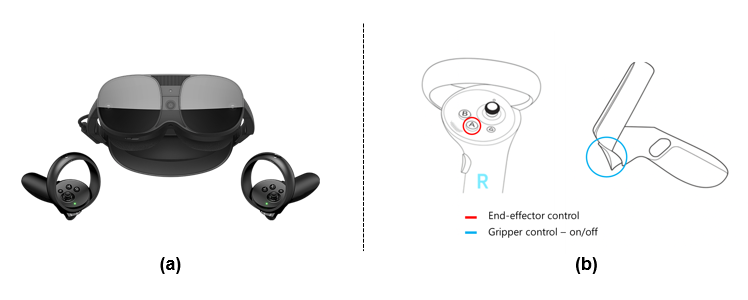}
    \caption{VR interface for teleoperation: (a) HTC VIVE XR Elite headset and controllers; (b) controller button mapping for end-effector control and gripper toggle}
    \label{fig5}
\end{figure}

The experimental platform is configured as a table-top manipulation system using a Franka Panda FR3 manipulator. All experiments are conducted within a bounded table-top workspace, and the same physical setup is used throughout the  obstacle-free, static-obstacle, and dynamic-obstacle scenarios. A VR headset with 6-DoF handheld controllers is used as the operator interface, as shown in Fig. ~\ref{fig5}(a). The controller pose serves as the primary input for teleoperation and is used to command end-effector motion in real time. As illustrated in Fig. ~\ref{fig5}(b), the VR controller inputs are assigned to two fundamental functions required for pick-and-place teleoperation: continuous end-effector control through the tracked controller motion and discrete gripper activation through a button-based on/off input. This input design provides an intuitive interaction scheme while keeping the command interface simple and consistent across all experiments.

\begin{figure}[!t]
    \centering
    \includegraphics[width=0.95\columnwidth]{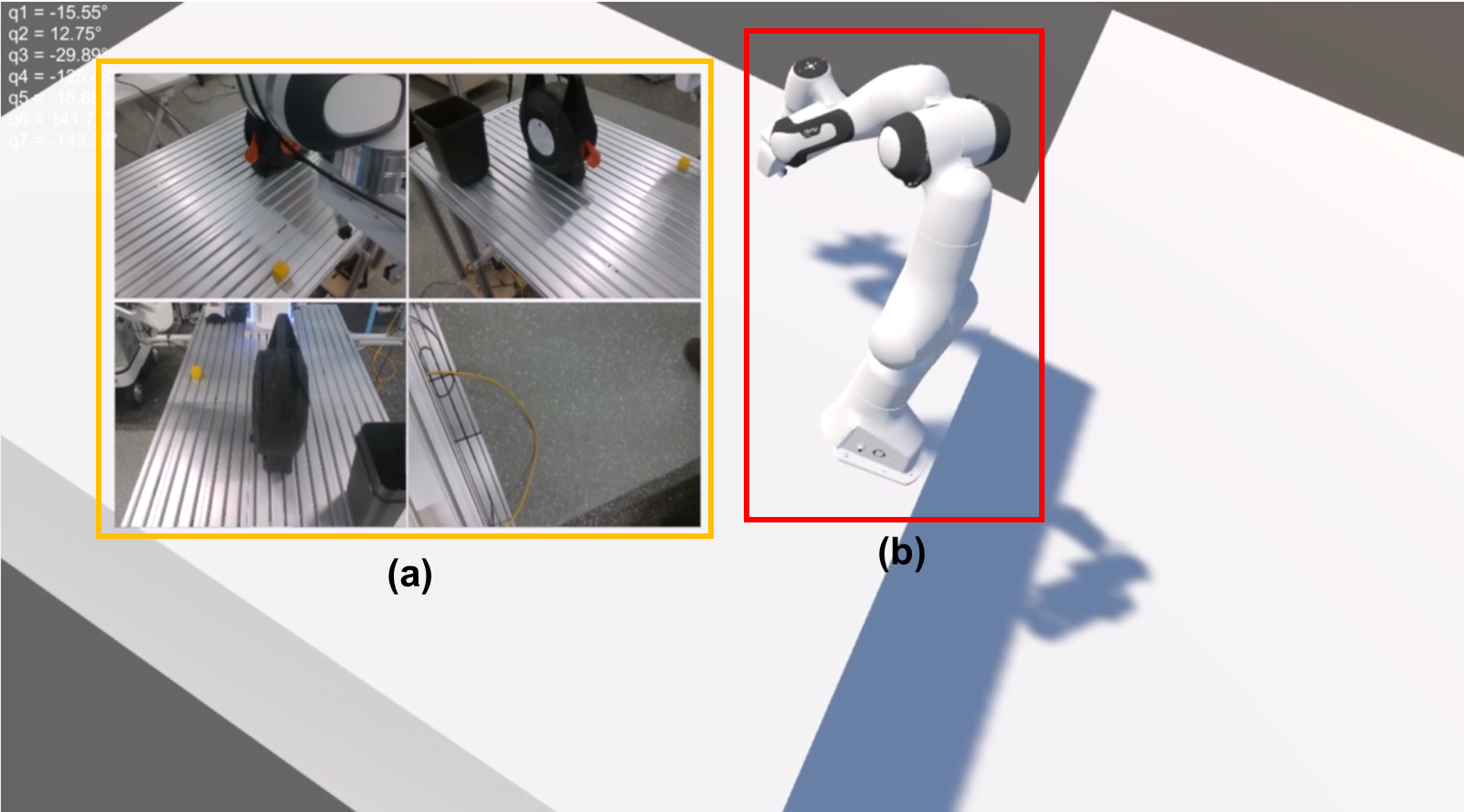}
    \caption{Hardware setup: (a) table-top manipulation setup with a Franka Panda FR3 and external/wrist RGB-D cameras; (b) reconstructed workspace volume used for perception and collision avoidance}
    \label{fig6}
\end{figure}

To enable dynamic environment perception, multiple RGB-D cameras are installed around the robot workspace, as depicted in Fig. ~\ref{fig6}(a). The cameras are placed to observe the table-top region from several viewpoints, allowing the system to capture both the robot motion and surrounding objects in the scene. The multi-view arrangement is particularly important for handling partial occlusions that can occur during manipulation, such as occlusion by the robot arm itself or by objects placed near the task region. The RGB-D observations from these cameras are used not only for operator visualization but also as the sensing input for online workspace reconstruction. Based on the sensing coverage of the installed RGB-D cameras, a bounded reconstruction region is defined around the robot’s operating area. Fig. ~\ref{fig6}(b) illustrates the 3D reconstruction workspace used in the experiments. This workspace volume is selected to include the main interaction region where pick-and-place actions take place and where obstacles are introduced in the static and dynamic settings. 

\subsubsection{Software Architecture and Communication}

The system software is organized to connect VR teleoperation, environment perception, and robot command generation within a unified runtime pipeline. The VR environment serves as the operator-facing front end, receiving user input and providing task feedback. As shown in Fig. 7(a), multi-view camera streams are displayed inside the VR environment to provide situational awareness of the remote workspace, allowing the operator to monitor robot motion, task objects, and obstacle conditions in real time. In addition, a virtual robot model synchronized with the real robot state is rendered in the VR scene (Fig. 7(b)), providing an interpretable visualization of robot posture and motion execution and helping the operator remain aware of configuration changes that may not be fully observable from a single camera viewpoint.

\begin{figure}[!t]
    \centering
    \includegraphics[width=0.95\columnwidth]{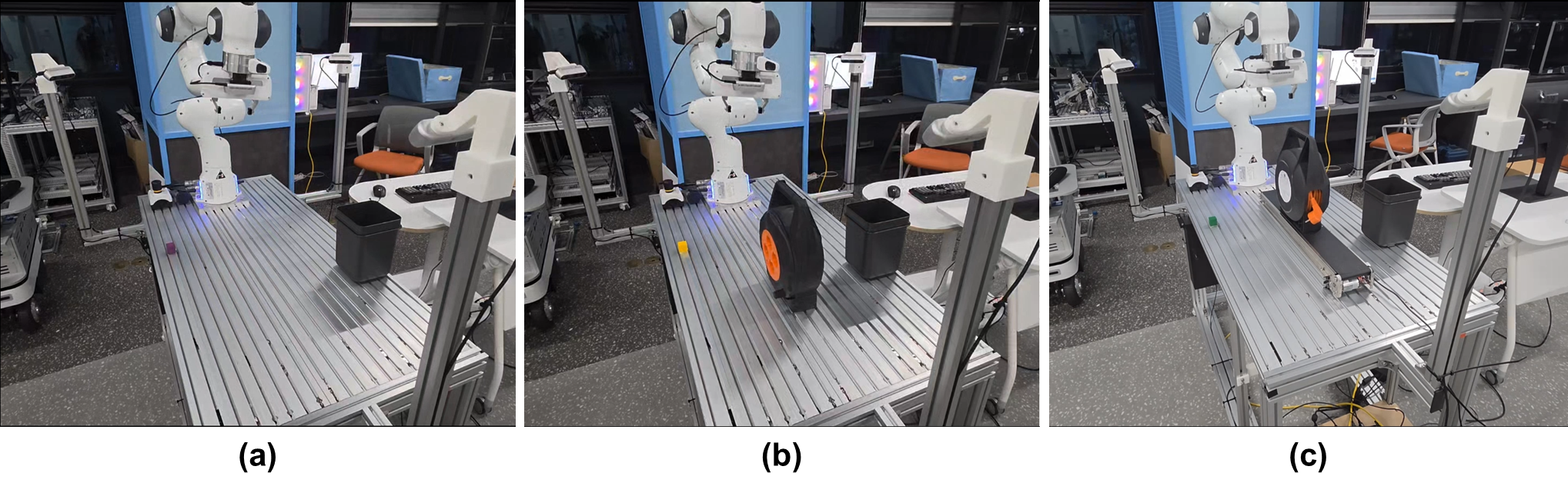}
    \caption{Unity-based teleoperation scene: (a) multi-view RGB-D streaming panel; (b) virtual robot model synchronized with the real robot state}
    \label{fig7}
\end{figure}

During teleoperation, the tracked 6-DoF controller motion is converted into a target end-effector pose and updated online. In parallel, the environment perception module processes RGB-D observations from the camera network to maintain an up-to-date representation of the workspace and extracts obstacle information for collision reasoning. This information is provided to the robot control module so that changes in the scene can be directly reflected in motion generation, enabling safe operation in the presence of obstacles.

In Experiment A (obstacle-free), the operator performs the task in a clear workspace, which serves as a baseline for evaluating tracking performance and computational stability. In Experiment B (static obstacle), a fixed obstacle is placed along the nominal motion path, and the operator is instructed to move the end-effector toward the obstacle region. This setup evaluates the system’s ability to generate collision-aware detours while maintaining stable motion, including cases with abrupt changes or pauses in operator input. In Experiment C (dynamic obstacle), the static obstacle is replaced with a moving object on a conveyor belt, resulting in continuously changing obstacle positions. This scenario evaluates the system’s ability to adapt its motion online to time-varying obstacles during both continuous manipulation and temporary pauses in operator input.
To satisfy the low-latency and high-reliability requirements of real-time teleoperation, the system adopts a dual communication architecture based on WebRTC \cite{c16} and ZeroMQ \cite{c17} (ZMQ). High-bandwidth visual data, including VR-rendered scenes and camera streams, are transmitted via WebRTC, which provides adaptive bitrate control and robust packet-loss handling for stable visual feedback. In contrast, time-critical control data such as target poses, robot states, and collision-related information are exchanged through ZMQ using lightweight low-latency communication patterns. This separation prevents visual streaming fluctuations from affecting command generation and robot motion. At runtime, the control module consumes the latest target pose and obstacle information to generate an executable joint command at each control step, enabling stable real-time teleoperation in dynamic environments where operator input and workspace conditions change simultaneously.

\subsection{Experimental Setup}

The proposed VR-based teleoperation framework is evaluated using a table-top pick-and-place task with a Franka Panda FR3 manipulator. Across all experiments, the operator performs the same task of grasping a block and placing it at a target location. The experimental scenarios differ in the presence and type of obstacles along the nominal motion path between the grasping and placement positions.

\begin{figure}[!t]
    \centering
    \includegraphics[width=0.95\columnwidth]{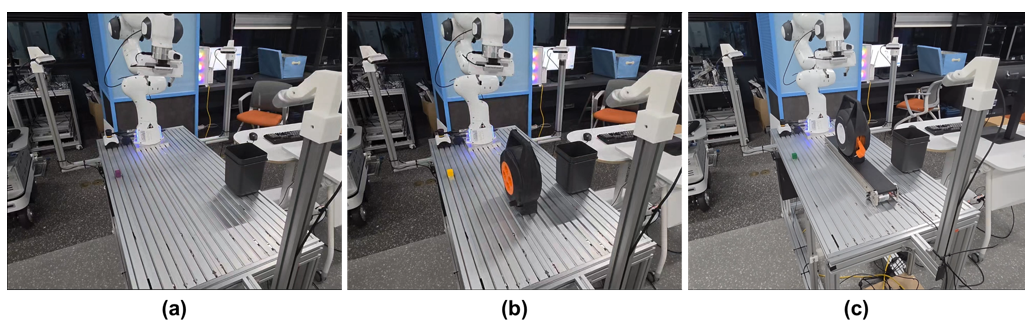}
    \caption{Experimental environments: (a) no obstacle; (b) static obstacle; (c) moving obstacle on a conveyor belt}
    \label{fig8}
\end{figure}

\begin{figure*}[!t]
    \centering
    \includegraphics[width=0.95\textwidth]{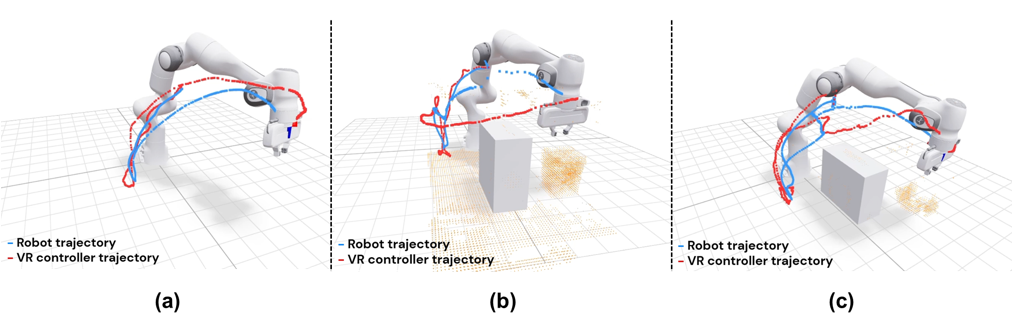}
    \caption{3D trajectories of the VR controller and the robot end-effector in different obstacle scenarios}
    \label{fig9}
\end{figure*}

As illustrated in Fig. ~\ref{fig8} and Fig. ~\ref{fig9}, three environments are considered: (A) an obstacle-free workspace, (B) a workspace with a static obstacle, and (C) a workspace with a dynamic obstacle. These scenarios are designed to evaluate real-time command generation and collision-aware motion behavior under progressively complex conditions.

In Experiment A (obstacle-free), the operator performs the task in a clear workspace, which serves as a baseline for evaluating tracking performance and computational stability. In Experiment B (static obstacle), a fixed obstacle is placed along the nominal motion path, and the operator is instructed to move the end-effector toward the obstacle region. This setup evaluates the system’s ability to generate collision-aware detours while maintaining stable motion, including cases with abrupt changes or pauses in operator input. In Experiment C (dynamic obstacle), the static obstacle is replaced with a moving object on a conveyor belt, resulting in continuously changing obstacle positions. This scenario evaluates the system’s ability to adapt its motion online to time-varying obstacles during both continuous manipulation and temporary pauses in operator input.

\subsection{Evaluation Metrics}

Two primary metrics are used to evaluate responsiveness and safety during teleoperation. Online planning latency is defined as the computation time required to generate an executable robot command at each control cycle. Specifically, the measured interval begins when the VR input is received and the optimization problem is formulated with the updated objective and constraints and ends when the final joint command for the corresponding control cycle is produced. This metric reflects the planning and command generation process and does not include sensing, communication, or low-level control execution delays. Collision avoidance performance evaluates whether safe motion is generated in obstacle-present environments while maintaining motion behavior consistent with the operator’s commanded task direction. In the static- and dynamic-obstacle scenarios, the evaluation focuses on whether the robot produces detour trajectories that satisfy collision constraints and avoids unsafe postures or abrupt motions. Behavior during operator pauses is also examined to confirm that the robot remains in a safe configuration rather than drifting toward collision-prone regions.

\section{Results}

\subsection{Online Planning Latency Results}

The online planning latency measured in the three environments is summarized in Table 1. For each experiment, the optimization time at each control cycle is recorded and the minimum, average, and maximum values are reported. In Experiment A (Fig ~\ref{fig10}), the average latency is about 20 ms, showing the fastest performance. In the presence of obstacles (Experiments B (Fig ~\ref{fig11}) and C (Fig ~\ref{fig12})), the average latency increases to roughly 35–38 ms, and the maximum latency exceeds 100 ms in segments where collision constraints are active. Thus, computation time increases with environmental complexity, especially when collision avoidance becomes critical. Nevertheless, the average latency remains within 30–40 ms, corresponding to real-time control at approximately 25–30 Hz.

\begin{table}[h]
    \centering
    \caption{Online planning latency across experimental settings}
    \label{tab1}
    \begin{tabular}{c|c|c|c}\hline\hline
    Experiment & Min (ms) & Avg. (ms) & Max (ms)\\\hline
    A & 14.377 & 20.178 & 62.809 \\
    B & 28.538 & 35.333 & 105.727 \\
    C & 30.117 & 38.309 & 110.595 \\\hline\hline    
    \end{tabular}
\end{table}

\subsection{Trajectory and Collision Avoidance Results}

\begin{figure}[!t]
    \centering
    \includegraphics[width=0.95\columnwidth]{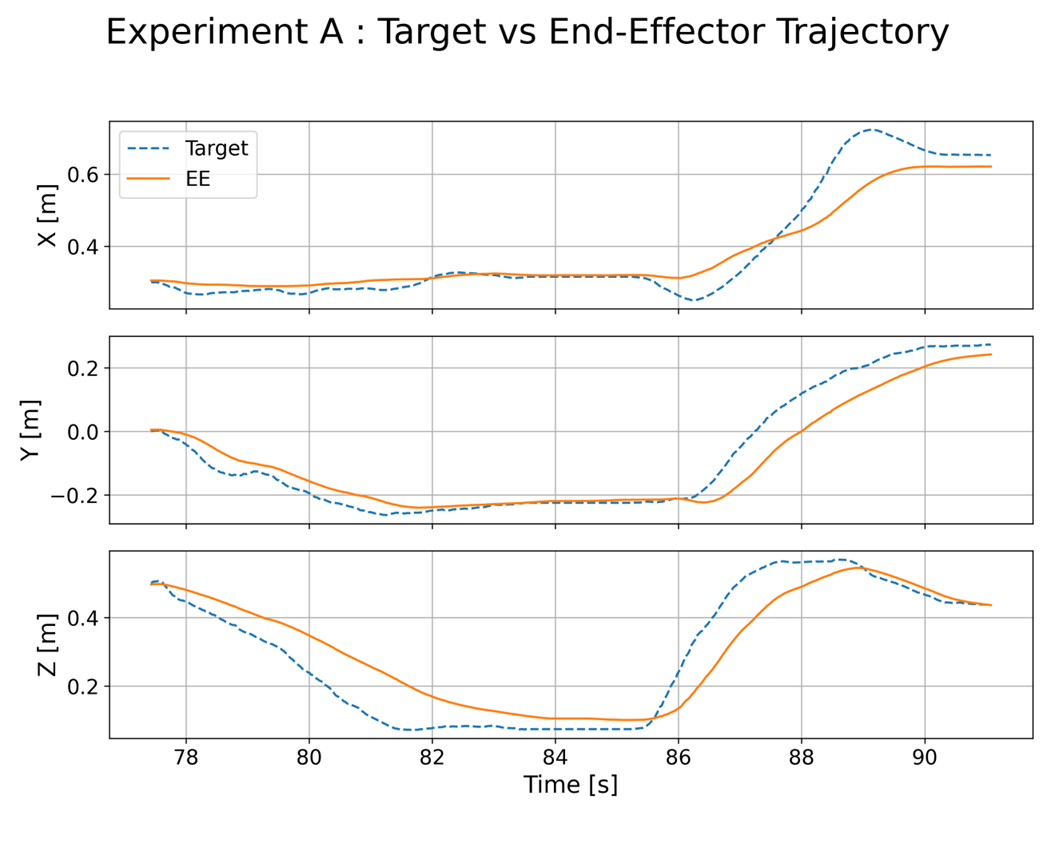}
    \caption{A: Axis-wise comparison between target and end-effector trajectories}
    \label{fig10}
\end{figure}

In the obstacle-free environment of Experiment A, the primary role of the optimizer is to accurately track the VR input, since no collision-related constraints are activated. Under this condition, the optimization problem is dominated by the target pose tracking term, allowing the solver to directly reflect the operator’s commanded motion. As shown in Fig. ~\ref{fig10}, the executed end-effector trajectory closely follows the VR controller command along all Cartesian axes, with only minor smoothing effects introduced by the trajectory optimization.

This indicates that the proposed control formulation does not introduce unnecessary distortion when additional safety constraints are absent. Consistently, the 3D trajectories shown in Fig. ~\ref{fig9}(a) reveal that the robot path almost completely overlaps with the controller path, demonstrating smooth and stable tracking behavior with minimal deviation throughout the task execution.

\begin{figure}[!t]
    \centering
    \includegraphics[width=0.95\columnwidth]{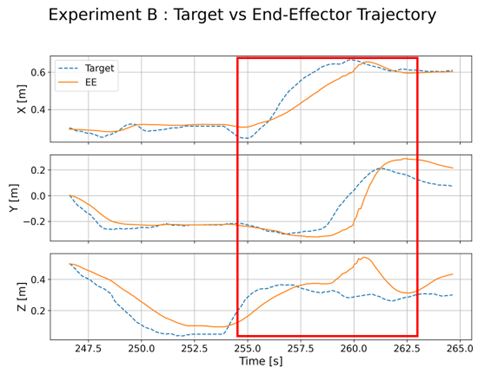}
    \caption{Experiment B: Axis-wise comparison between target and end-effector trajectories}
    \label{fig11}
\end{figure}

The results for Experiment B, in which a static obstacle is placed along the task path, are presented in Fig. ~\ref{fig11}. The red-boxed intervals in Fig. ~\ref{fig11} indicate periods during which collision-avoidance constraints become active as the robot approaches the obstacle. While the VR controller command follows a nearly straight path that intersects the obstacle region, the executed end-effector trajectory deviates from the commanded path to satisfy the imposed collision constraints. This behavior highlights the role of the optimizer in reshaping the trajectory online rather than strictly following the input command. As illustrated in Fig. ~\ref{fig9}(b), the robot trajectory bends smoothly around both the physical obstacle and the reconstructed point cloud representation, forming a safe detour while still converging toward the user-specified target region. These results indicate that the proposed framework maintains motion behavior consistent with the operator’s commanded task direction, even when local deviations are required to ensure collision-free motion.

\begin{figure}[!t]
    \centering
    \includegraphics[width=0.95\columnwidth]{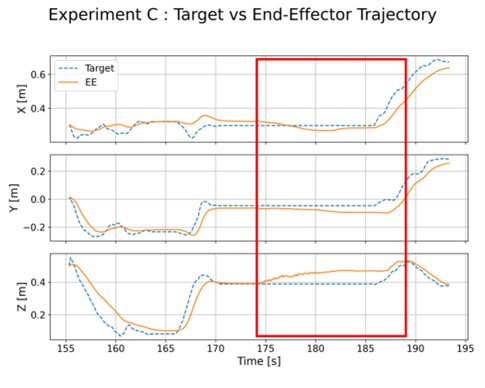}
    \caption{Experiment C: Axis-wise comparison between target and end-effector trajectories}
    \label{fig12}
\end{figure}

Experiment C further evaluates the system performance in the presence of a dynamic obstacle moving on a conveyor belt, with the corresponding results shown in Fig. ~\ref{fig12}. The highlighted intervals in Fig. ~\ref{fig12} represent moments when the obstacle approaches the robot and collision-related constraints are activated dynamically. During these intervals, the end-effector trajectory temporarily departs from the VR controller command to maintain a safe separation distance from the moving obstacle. Once the obstacle moves away and the collision constraints are relaxed, the trajectory gradually returns toward the commanded path. The 3D visualization in Fig. ~\ref{fig9}(c) clearly illustrates how the robot trajectory is continuously reshaped in response to the obstacle motion, while the controller trajectory directly reflects the user’s hand movement without modification. Notably, when the operator pauses manipulation, the robot transitions to a safer configuration instead of remaining near the obstacle, and subsequently resumes motion toward the operator-commanded task direction after the obstacle has passed.

Overall, these experimental results indicate that the proposed framework achieves accurate and stable tracking of user inputs in obstacle-free conditions, while balancing task tracking and safety constraints in environments with static and dynamic obstacles. By integrating real-time perception with GPU-based trajectory optimization, the system rapidly generates feasible detour trajectories that respond to environmental changes online while remaining consistent with the operator’s commanded motion under safety and real-time constraints.

\section{conclusion}
This study presented a VR-based teleoperation framework for achieving stable robot manipulation in dynamic environments and demonstrated its feasibility on a real Franka manipulator. The proposed system translates the operator’s hand motion in a VR interface into end-effector commands in real time, while simultaneously incorporating 3D environmental information to support collision-aware motion generation. As a result, the system enables safe and responsive robot behavior under changing environmental conditions.
 
The main contributions of this work can be summarized as follows. First, multiple control objectives—including target pose tracking, joint-limit avoidance, collision avoidance, and posture regularization—are formulated within a single optimization problem. This unified formulation enables motion generation that remains consistent with the operator’s commanded motion while adapting to environmental constraints, providing a more structured way to account for environmental constraints than teleoperation approaches based solely on simple inverse kinematics. Second, by leveraging a GPU-based optimization solver, the system achieves online planning latencies of approximately 30–40 ms (corresponding to an update rate of about 25–30 Hz), allowing VR inputs to be reflected in robot motion with minimal delay. Third, the environment perception module, combining TSDF-based 3D reconstruction (NvBlox), robot occupancy filtering, and DBSCAN-based clustering, continuously updates obstacle information and integrates it directly into the collision cost term, enabling prompt responses to dynamic changes in the workspace.
Experimental results in both static- and dynamic-obstacle scenarios demonstrate that the proposed framework can generate collision-aware trajectories while maintaining motion behavior consistent with the operator’s commanded task direction. These results indicate that the proposed approach can serve as a practical foundation for real-time teleoperation in environments where direct human intervention is difficult or unsafe.

In addition, further comparative evaluation with conventional IK-based and optimization-based teleoperation approaches would help more clearly characterize the relative properties and applicability of the proposed framework. Future work will focus on extending the framework to mobile manipulators and more complex real-world environments, including longer-term validation under diverse task conditions.

\bibliographystyle{IEEEtran}
\bibliography{bibtex}

\end{document}